\documentclass[conference]{IEEEtran}
\IEEEoverridecommandlockouts
\usepackage{cite}
\usepackage{amsmath,amssymb,amsfonts}
\usepackage{algorithmic}
\usepackage{graphicx}
\usepackage{textcomp}
\usepackage{xcolor}
\usepackage[ruled,vlined]{algorithm2e}
\def\BibTeX{{\rm B\kern-.05em{\sc i\kern-.025em b}\kern-.08em
    T\kern-.1667em\lower.7ex\hbox{E}\kern-.125emX}}
\begin{document}

\title{Partial Hierarchical Pose Graph Optimization for SLAM\\
}

\author{\IEEEauthorblockN{Alexander Korovko}
\IEEEauthorblockA{
\textit{NVIDIA}\\
Moscow, akorovko@nvidia.com}
\and
\IEEEauthorblockN{Dmitry Robustov}
\IEEEauthorblockA{
\textit{NVIDIA}\\
Moscow, drobustov@nvidia.com}
}

\maketitle

\begin{abstract}
In this paper we consider a hierarchical pose graph optimization (HPGO) for Simultaneous Localization and Mapping (SLAM).
We propose a fast incremental procedure for building hierarchy levels in pose graphs. We study the properties of this procedure and show that our solution delivers high execution speed, high reduction rate and good flexibility.
We propose a way to do partial hierarchical optimization and compare it to other optimization modes. We show that given a comparatively large amount of poses, partial HPGO gives a 10x speed up comparing to the original optimization, not sacrificing the quality.
\end{abstract}


\section{Introduction}
SLAM is a method of estimating robot position in an unknown environment along with the simultaneous construction the 3D map of this environment. There exist several different approaches of solving of this problem, e.g. particle filters, Kalman filters, graph-based SLAM etc.  Amongst them, the graph-based
approach gained much popularity in the last decade thanks to its efficiency and flexibility.\newline
\\
Graph-based method considers the trajectory of the robot as a directed graph. The nodes of such graph correspond to estimations of the respective robot positions. The edges correspond to robot`s observations that come from odometry.\\
\\
Graph-based SLAM usually consist of two main parts. First one is the construction of the graph according to odometry results. This is also known as a frontend part. The second part is known as SLAM backend. It aims to find the most likely robot positions given robot observations. It has been shown, that such problem can be formulated as a minimization of a squared-error function problem and can be effectively solved using well known optimizers like Gauss-Newton, Levenberg–Marquardt etc. \\\\
In this paper, we focus on SLAM backend and propose an effective and flexible way to reduce the computation complexity of the pose graph optimization. We suppose, that to achieve high accuracy one doesn`t need to perform the optimization on the whole set of graph nodes. More beneficial is to select a subset of nodes for that purpose and do the optimization only on them. To further improve our results we combine this partial optimization with hierarchical pose graph approach. We also propose a way to construct hierarchy levels in pose graph, which allows fast addition of new nodes into hierarchy structure, without the need of rebuilding it.

\section{Related work}
SLAM graph optimization have been studied intensively in the past. The least square error minimization approach for finding the maximum likelihood maps was first proposed by Lu and Milios \cite{b1}, though their solution was considered as computationally expensive. 
Later Gutmann and Konolige \cite{b2} proposed a way of graph construction that allows detection of loop closures while performing an incremental estimation algorithm. Howard \cite{b3} proposed to apply relaxation to build a map. Duckett \cite{b4}  - to use Gauss-Seidel relaxation to minimize the error in pose graph. Frese \cite{b5} introduced multi-level relaxation, which provides remarkable results in flat environments especially if the error in the initial guess is limited. Olson \cite{b6} presented an efficient optimization approach which is based on the stochastic gradient descent and can efficiently correct even large pose graphs. Grisetti \cite{b7} proposed an extension of Olson’s approach that uses a tree parameterization of the nodes in 2D and 3D to speed up the convergence. \\\\
Grisetti and Stachniss in \cite{b8} introduced a hierarchical pose-graph optimization and showed that one doesn`t need to perform the optimization over the whole set of graph nodes. More beneficial in the sense of speed/accuracy trade-off is to group the nodes, that lie close to each other, in sets. Select the representative node in such sets and do optimization over such representative node only. Once the optimization is done, each representative node shares it`s computed transformation with the nodes of it`s group.

\section{Construction of the Hierarchy}

We start from considering a problem of hierarchy construction in pose graphs, i.e. a way to split the nodes in graph into groups.\\

For this purpose we propose to use a papular maximum modularity criteria \cite{b9}. The modularity of a partition is a scalar value between $-1$ and $1$
that measures the density of links inside groups as compared to links between
groups.
\begin{equation}
     Q = \frac{1}{2m}\sum_{i, j}^{}\left [ A_{ij} -\frac{k_ik_j}{2m} \right ]\delta (g_i, g_j),
\end{equation}
where $A_{ij}$ represents the weight of the edge between nodes $i$ and $j$, $k_i = \sum_{j}A_{ij}$ is the sum of the weights of the edges attached to node $i$, $g_i$
is the group to which node $i$ is
assigned, the $\delta$-function $\delta(u, v)$ is $1$ if $u = v$ and $0$ otherwise and $m = \frac{1}{2} \sum_{ij}A_{ij}$.
\\\\
As an edge weight here we use an inverse Mehalanobis distance:
\begin{equation}
     W_{i,j} = \frac{1}{z_{i,j}^T\Omega_{i,j} z_{i,j}},
\end{equation}
where
\begin{equation}
     \boldsymbol{z} = ({x}', {y}', {z}', {\rho}' , {\varphi}' , {\theta}' )^{T}
\end{equation}
\begin{equation}
     \boldsymbol{\Omega} = R^{6\textrm{x}6}
\end{equation}
are the mean and the covariance matrix of robot observations respectively. Here $z$ is represented as an element $SE(3)$ and encodes the translation and rotation in 3D, i.e twist.
\\\\
Direct optimization of modularity \cite{b10} is proved to be NP-hard and hence a set of heuristic ways were proposed to do such optimization \cite{b11}, \cite{b12}, \cite{b13}. Among them the Louvain algorithm \cite{b14} seems to us very convenient for our application, due to it`s high speed any ease of computations. 
\begin{figure*}[h]
\includegraphics[width=\textwidth]{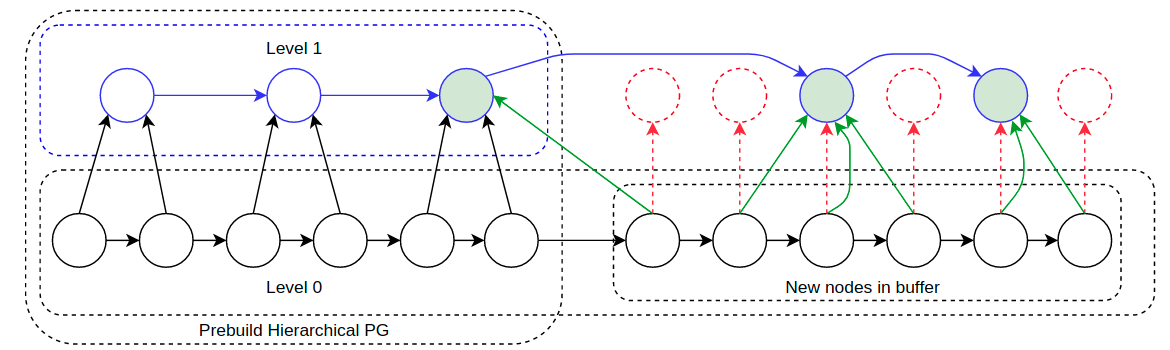}
\caption{Illustration of Algorithm 1. Black nodes are the nodes of the first level. They form two sets denoted $N$ and ${N}'$. Red nodes correspond to initially assigned groups to nodes of ${N}'$. Green nodes correspond to finally created/updated groups.}
\label{fig}
\end{figure*}

For every node $i$ this algorithm considers the neighbours $j$ of $i$ and evaluates
the gain of modularity that would take place by removing $i$ from its group and by
placing it in the group of $j$. The node $i$ is then placed in the group for which this gain is maximum and positive.
\begin{equation}
     \Delta Q = \left [ \frac{\Sigma _{in}}{2 * m} - \frac{\Sigma _{group} * k _{node}}{2 * m^2} \right ],
\end{equation}
where $\Sigma _{in}$ is the total weight of edges inside neighbour group incident with the current node, $\Sigma _{group}$ - total weight of edges related to a neighbour group, $k _{node}$ - total weight of edges incident with the current node.\\


We focus on building the algorithm that allows to construct an additive hierarchy, i.e. to add newly created nodes into already existing groups without the need of all groups rebuilding. We assume that before the start we have a set of graph nodes $n \in N$. Every element of this set is assigned to a group $g \in G$ with a mapping $M: N\rightarrow G$ . Along with set $N$ there is also a set ${N}'$ of nodes, that are not assigned to any group. Our aim is to find a mapping:
\begin{equation}
     {M}': N\cup {N}'\rightarrow G\cup {G}'
\end{equation}

We start with assigning each node ${n}' \in {N}'$ to it`s own new group ${g}' \in {G}'$. We iterate over these nodes, calculating the modularity gain from moving this node into it`s neighbours group. We select maximum of the gain over neighbours. If the maximum gain is positive, we move the node to the respective group.\\\\
Due to iterative nature, the algorithm may yield arbitrarily badly connected groups. By moving some node into it`s neighbour`s group at one iteration we may move the neighbour away of his own group on the next iteration. To deal with this problem we simply prohibit assigning the node to any group in case it`s current group has more than one element.\\\\
To make the coverage of nodes by groups more uniform, we add additional limitation on the group size: we limit it with some chosen value, $3$ in our experiments.\\
\\
The pseudo code of our algorithm is presented below.

\begin{algorithm}[h]
\caption{$BuildHierarchy$}
\SetAlgoLined
 \textbf{Input:} $N$ - set of nodes, $G$ - set of groups, $M: N\rightarrow G$ - mapping between nodes and groups, ${N}'$ - set of unassigned nodes\;
 \textbf{Initialization:} Assign each $n \in {N}'$ its own group\;
 \For{node in ${N}'$}
 {
   \If{$GroupSize(node) > 1$}{
      continue\;
   }
   max\_gain = -1\;
   top\_neighbour = None\;
   \For{neighbour in GetNeighbours(node)}{
      gain = $\Delta Q(node, neighbour)$\;
      neighbour\_group\_size = GroupSize(neighbour)\;
      \If{$neighbour\_group\_size \leq K$}{
          \If{$gain > max\_gain$}{
             max\_gain = gain\;
             top\_neighbour = neighbour\;
          }
      }
      \If{$max\_gain > 0$}{
         //Assign node to top\_neighbour`s group\;
         ${M}'[node] =$ top\_neighbour`s group\;
      }
   }
 }
\end{algorithm}

Groups, produced by our algorithm, form the set of nodes for the level-2 pose graph. The edges of this graph can be obtained, as it was proposed in \cite{b8}.
In our experiment we use Algorithm 1 to build new hierarchy levels sequentially, when the size of the current top level exceeds some predefined threshold, 300 nodes in our experiments.

\begin{figure}[h]
\centerline{\includegraphics[width=8cm]{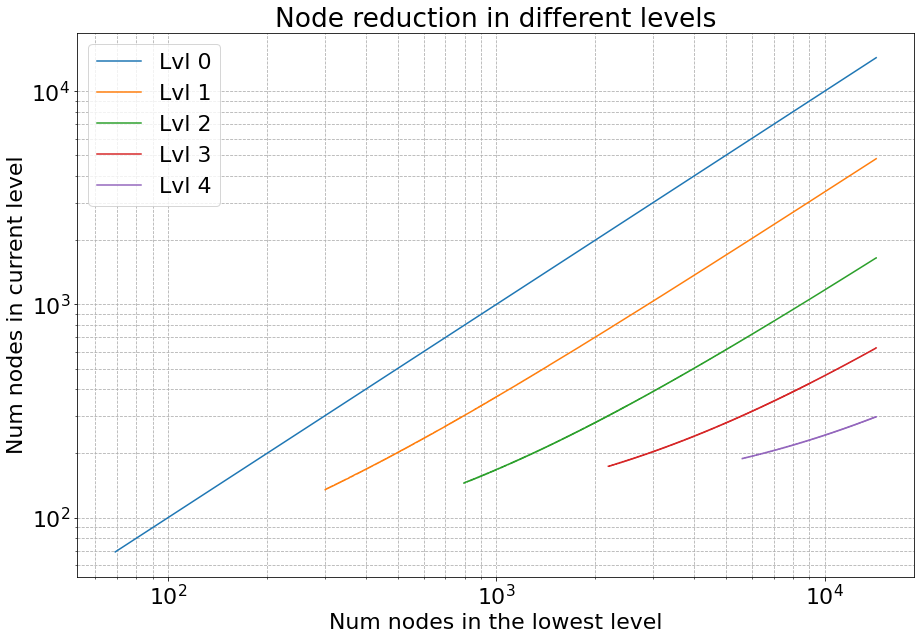}}
\caption{Size (number of nodes) of hierarchy levels relative to the original pose graph size. Our algorithm shows the reduction rate almost 3x times between subsequent pose graph levels.}
\label{fig}
\end{figure}

\newpage

\section{Partial Hierarchical Optimization}
In this part we propose a way of performing the partial optimization on the hierarchical pose graph containing several levels of hierarchy. 

\subsection{Pose graph optimization on manifold}
The goal of original pose graph optimization, as described in \cite{b1}, is to find the most likely construction of pose graph nodes with respect to given observations. Let $\boldsymbol{x} = (x_1, x_2,  ..., x_n)^T$ be a state vector of graph, where $x_i$ are the states of each node respectively. Let also $z_{ij}$ and $\Omega_{ij}$ be the mean and covariance matrix of the observation on node $x_j$ from node $x_i$. Given this notions we can define a function $e_{ij} = f(x_i, x_j, z_{i,j})$, which measures the difference between the expected observation $x_j$ from $x_i$ and the measurement $z_{ij}$.\\
Let V be the set of pose graph edges. Then the goal of pose graph optimization is to find such $\boldsymbol{x^*}$ that minimizes the negative log-likelihood function of all the observations.
\begin{equation}
    F(x) = \sum_{i, j \in V} e_{ij}\Omega_{ij}e_{ij}^T
\end{equation}
\begin{equation}
     \boldsymbol{x^*} = argmin_x F(x)
\end{equation}

The minimization in equation (8) is usually done with a Gauss-Newton or Levenberg–Marquardt approaches using sparse Cholesky factorization, which reduce the task to iterative solution of a linear equation and updating the estimated $x$.

\begin{equation}
    H \Delta x = - b
\end{equation}
\begin{equation}
     x_k =  x_{k - 1} + \Delta x
\end{equation}
where
\begin{equation}
    H = \sum_{i, j \in V} J_{ij}^T \Omega_{ij} J_{ij} 
\end{equation}
\begin{equation}
    b = \sum_{i, j \in V} e_{ij}^T \Omega_{ij} J_{ij}
\end{equation}
where 
\begin{equation}
   J_{ij} = \left ( \frac{\partial e_{ij}(x + \Delta x)}{ \partial \Delta x} \right ) _{x = x_{k-1}}
\end{equation}

Let $|N|$ and $|V|$ be the number of nodes and edges in the pose graph respectively.
Equations 11 and 12 involve the summation of the terms that include jacobians over all the graph edges.
The time complexity of this summation is $O(|V|)$. Linear equation 9 is solved using sparse Cholesky factorization.
The time complexity of the sparse Cholesky factorization depends on the exact size and structure of matrix $H$. For simplicity, we can consider it as $O(|N|)$ as matrix H is usually a highly sparse matrix. Such assumption is confirmed by our experiments (Fig. 4, green line). In this way, the overall time complexity of pose graph optimization is $O(|V| + |N|)$.

\subsection{Partial hierarchical pose graph optimization}
Partial hierarchical pose graph optimization (PHPGO) relies on a special structure of the hierarchical levels in pose graph. It assumes that there exist an arbitrary number of hierarchy levels. A new level is built when the number of nodes in the top level exceeds some predefined threshold. Considering these features, doing the optimization only on the top-level of hierarchy may lead to significant degradation of the optimization quality due to the loss of information.
\\\\
Instead we propose to do the optimization sequentially on all hierarchy levels. On each level we select a subgraph consisting of a certain number of nodes and do optimization only in this subgraph. \\\\
First we start with the full optimization on the top level and apply obtained transformations to the nodes of the subsequent level. From level $N - 1$ and on we use a breads-first search to find $P$ nodes in graph closest to the last added node and do the optimization only on the selected nodes. Nodes lying on the border of resulting set serve as a constrained nodes.
\\\\
\begin{figure}[h]
\centerline{\includegraphics[width=8cm]{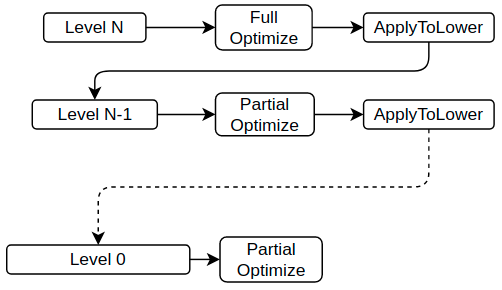}}
\caption{Partial hierarchical optimization diagram.}
\label{fig}
\end{figure}

To estimate the time complexity of the proposed procedure let`s denote $K$ to be a ratio between number of nodes in two subsequent hierarchy levels. In out experiments we used $K = 3$. Using this and the criteria of building new level, we can estimate the total number of hierarchy levels stored in memory:
\begin{equation}
   L = log_K\left ( \frac{|N|}{T} \right )
\end{equation}
where $T$ is the threshold used to build a new level. $T = 300$ in out experiments.
The optimization on the each level takes constant time due to the number of nodes involved in optimization is constant. Hence the time complexity of partial hierarchical optimization is $O(log_K\left ( \frac{|N|}{T} \right ))$. Experimental results are provided on Fig. 4.\\

\begin{figure}[h]
\centerline{\includegraphics[width=8cm]{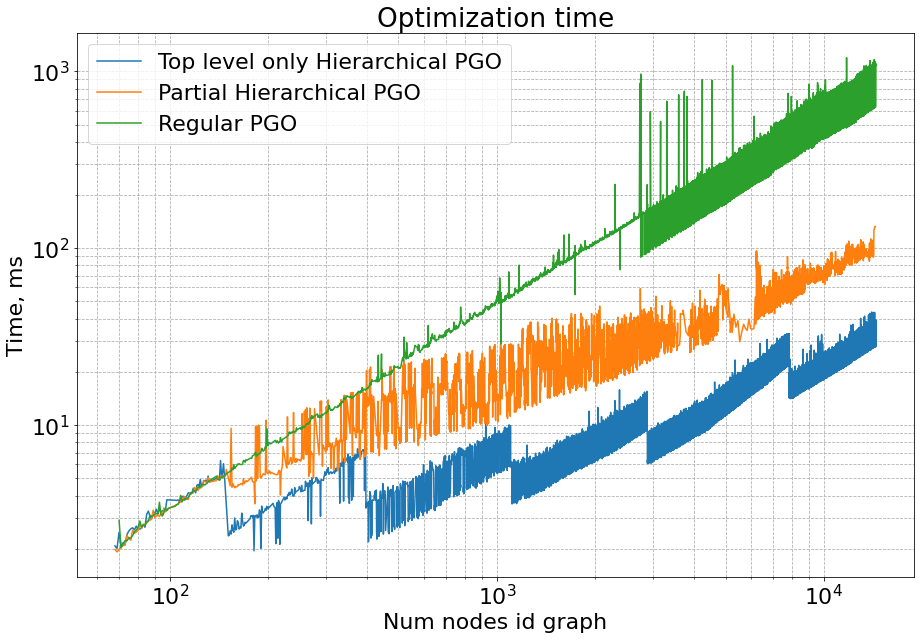}}
\caption{Inference time of partial HPGO(orange) relative to the original PGO(green) and optimization of the highest level only(blue).}
\label{fig}
\end{figure}

To show, that our algorithm, though being very fast, doesn`t suffer from quality degradation we provide the metrics calculated both for regular pose graph optimization and for PHPGO on Kitti dataset \cite{b16}. We used two commonly used metrics: ATE - average translation error and ARE - average rotation error. These metrics are presented in Table 1.
\\

\begin{table}
\centering
\caption{SLAM metrics on Kitti sequences.}
\label{table:1}
\renewcommand{\arraystretch}{2}
\begin{tabular}{ |c|c|c|c|c| } 
\hline
\textbf{Sequence №} & \multicolumn{2}{|c|}{\textbf{Default PGO}} & \multicolumn{2}{|c|}{\textbf{Partial HPGO}} \\
\hline
-- & ATE & ARE & ATE & ARE\\
\hline
Kitti-00 & 0.00731255 & 0.00304609 & \textbf{0.00696196} & \textbf{0.00224263} \\
\hline
Kitti-01 & 0.02225725 & 0.00240395 & 0.02225725 & 0.00240395 \\
\hline
Kitti-02 & \textbf{0.00670220} & 0.00179047 & 0.00681975 & \textbf{0.00177204} \\
\hline
Kitti-03 & 0.00933016 & 0.00220685 & 0.00933016 & 0.00220685 \\
\hline
Kitti-04 & 0.00879484 & 0.00180973 & 0.00879484 & 0.00180973 \\
\hline
Kitti-05 & \textbf{0.00492391} & 0.00156959 & 0.00493906 & \textbf{0.00155226} \\
\hline
Kitti-06 & 0.00590030 & \textbf{0.00123359} & \textbf{0.00588521} & 0.00124767 \\
\hline
Kitti-07 & 0.00485413 & 0.00255782 & \textbf{0.00480950} & \textbf{0.00254468} \\
\hline
Kitti-08 & 0.00974305 & 0.00271669 & 0.00974305 & 0.00271669 \\
\hline
Kitti-09 & 0.00919602 & \textbf{0.00199114} & \textbf{0.00919584} & 0.00199118 \\
\hline
Kitti-10 & 0.00573968 & 0.00217407 & 0.00573968 & 0.00217407 \\ 
\hline
\hline
\textbf{Average} & 0.00802947 & 0.00227903 & \textbf{0.00797863} & \textbf{0.00209171} \\ 
\hline
\end{tabular}
\end{table}

\section{Results}

We proposed an effective way to build hierarchy structure for pose graphs based on the modularity maximization. A big advantage of this algorithm is that it supports sequential addition of newly created nodes into the hierarchy structure. 
Proposed partial hierarchical optimization is proved to be very effective from the computational point of view and is much faster comparing to the original pose graph optimization. We showed that our approach doesn`t lead to the degradation of metrics, moreover, even improves the average metrics on Kitti benchmark.

\end{document}